\documentclass[10pt,twocolumn,letterpaper]{article}

\usepackage{cvpr}             

\usepackage{microtype}
\usepackage{multirow}
\usepackage{makecell}
\usepackage{enumitem}
\usepackage{caption}
\usepackage{subcaption}
\usepackage{xcolor}
\newcommand{\supp}[1]{Supp.~#1}
\newcommand{\ALL}{\textsc{all}}
\newcommand{\RAND}{\textsc{rand}}
\newcommand{\AGREE}{\textsc{agree}}
\newcommand{\SELECT}{\textsc{select}}
\newcommand{\SELECTR}{\textsc{select+repair}}
\newcommand{\PARENT}{\textsc{select+parent}}
\newcommand{\RANDR}{\textsc{select+random-repair}}

\newcommand{\BASE}{\textsc{base}}
\newcommand{\dsel}{\Delta_{\mathrm{select}}}
\newcommand{\drep}{\Delta_{\mathrm{repair}}}

\newcommand{\RANDG}{\textsc{select+random-region}}
\newcommand{\RANDM}{\textsc{rand-matched}}

\definecolor{cvprblue}{rgb}{0.21,0.49,0.74}

\usepackage{colortbl}
\definecolor{rowhl}{RGB}{232,245,240}
\newcolumntype{L}[1]{>{\raggedright\arraybackslash}p{#1}}
\newcolumntype{C}[1]{>{\centering\arraybackslash}p{#1}}
\SetTracking{encoding=*}{-8}
\DeclareRobustCommand{\tightpara}{\lsstyle\looseness=-1}
\newcommand{\tablefont}{\fontsize{8.5}{10}\selectfont}
\usepackage[pagebackref,breaklinks,colorlinks,allcolors=cvprblue]{hyperref}
\usepackage[capitalize]{cleveref}
\crefname{section}{Sec.}{Secs.}
\Crefname{section}{Section}{Sections}
\Crefname{table}{Table}{Tables}
\crefname{table}{Tab.}{Tabs.}
\crefname{equation}{Eq.}{Eqs.}
\Crefname{equation}{Equation}{Equations}

\def\confName{CVPR}
\def\confYear{2027}

\title{Task-Conditioned Detail Allocation for Synthetic Segmentation Data}

\author{Chunming He$^{1}$\,,
Dingming Zhang$^{1}$\,,
Jifan Zhang$^{1}$\,,
Yuhan Dong$^{1}$\,,
Longxiang Tang$^{2}$\,, \\
Youwei Pang$^{3}$\,, 
Xiaoqi Zhao$^{3}$\,, 
            Fengyang Xiao$^{1,\dagger}$\,, 
        and {Sina Farsiu}$^{1,\dagger}$\\
        $^1$Duke University,
	$^2$Harvard University,  $^3$Nanyang Technological University 
\\
$\dagger$ Corresponding Author, \quad
Contact: chunming.he@duke.edu
}

\patchcmd{\abstract}{\vspace*{12pt}}{\vspace*{6pt}}{}{}

\begin{document}
\twocolumn[{
\maketitle
\begin{center}
\includegraphics[width=\textwidth]{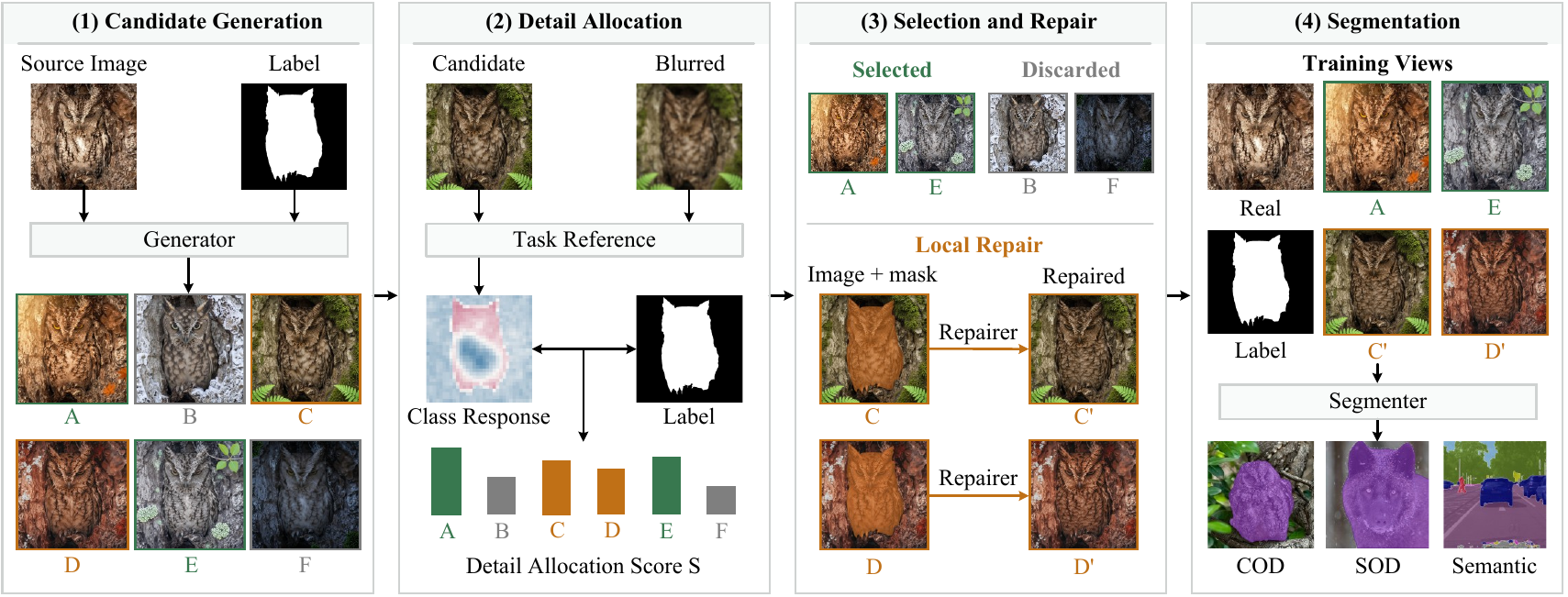}\vspace{-1mm}
\captionof{figure}{\textbf{Selection and local repair of synthetic segmentation data.}
(1) Six candidates are generated from a source image and its label.
(2) A task reference scores the spatial alignment of blur-induced class responses with the label.
(3) Up to two candidates are selected, and up to two others are considered for repair. Source-relative region contributions trigger whole-foreground edits in binary tasks and guide region selection in multiclass tasks.
(4) Selected candidates and accepted repairs reuse the label and train the segmenter with real images.}
\label{fig:teaser}
\end{center}
}]

\begin{abstract}
\lsstyle
Synthetic images sharing a segmentation label can differ in their training value. We propose Task-Conditioned Detail Allocation (TCDA) to select useful candidates and locally edit some remaining images. We ask whether image details provide class-specific cues in the regions marked by the label. We train a small task reference on frozen image features from real training images. We compare its class responses before and after blurring each candidate, and score how well these response changes align with the labeled regions. This detail allocation score ranks candidates from the same source. Comparing their response--label alignment with that of the real source image also guides repair decisions and, in multiclass tasks, the choice of regions to edit. Selected candidates and accepted repairs are used for training with the original labels and the unchanged downstream segmenter. Experiments on camouflaged object detection, degraded salient object detection, and adverse-condition semantic segmentation show that our selection outperforms equally sized random and label-agreement subsets. Combining selection and repair improves over real-only training by up to 0.019 in $S_\alpha$ on COD10K and 2.86 mIoU points on ACDC.
\mbox{Code and the synthetic pools will be released.}
\par
\end{abstract}

\setlength{\abovedisplayskip}{2pt}
\setlength{\belowdisplayskip}{2pt}

\section{Introduction}
\label{sec:intro}

Segmentation datasets often lack labeled images covering the conditions encountered in practice. Driving datasets underrepresent night, fog, and rain, degraded salient object images are scarce, and camouflaged object detection (COD) has only a few thousand densely annotated training images.

Generative models can create training images from existing annotations by conditioning on a label map or editing its source photograph \cite{rombach2022ldm,zhang2023controlnet,jia2024dginstyle,zhao2024lakered}. We study candidates that share a label and are intended to preserve object layout while varying lighting, texture, and surroundings. After excluding known image-label conflicts, the remaining candidates can still differ considerably in training utility.

Image quality, image-text agreement, label consistency, and instance fidelity provide useful measures for evaluating synthetic training images \cite{ke2021musiq,radford2021clip,yang2023freemask,tang2025sds,zhang2026sense}. However, agreement between a model's prediction and a label does not reveal how image details contribute to that prediction. This distinction matters for camouflaged objects: making an outline more visible does not necessarily make a generated image more representative of real camouflage. We therefore ask whether image details provide class-specific cues in the regions marked by the label. We call this spatial relationship \emph{detail allocation} and examine whether it helps identify useful synthetic training images for downstream segmentation.

Our method, TCDA, measures this relationship using a small task reference trained on frozen DINOv2 features \cite{oquab2024dinov2} from real training images. For each candidate, we compare the reference's class responses before and after Gaussian blurring. The spatial correlation of these response changes with the labeled regions gives the detail allocation score (\cref{fig:teaser}), which ranks candidates from the same source. We select up to two of six candidates, favoring views distinct from the source and each other when scores are nearly tied.

We also consider up to two unselected candidates per source for repair. Region contributions to the score difference from the real source determine whether to edit a candidate. Binary tasks edit the complete foreground; multiclass tasks also use these contributions to choose complete label regions. The task's generator proposes edits, and accepted repairs retain the inherited labels and join the selected training views alongside the real training images.

Our contributions are summarized as follows:
\begin{itemize}[itemsep=1pt,topsep=2pt]
\item We introduce a task-conditioned detail allocation score that measures how blur-induced class responses align with segmentation labels, and use it to select synthetic images.
\item We derive source-relative region contributions from the same score to trigger repair of the complete foreground in binary tasks and guide region selection in multiclass tasks.
\item We evaluate selection and repair across three segmentation tasks, with COD comparisons using matched candidate sets, unedited parents, and random-location repair.
\end{itemize}

\section{Related Work}
\label{sec:related}

\noindent\textbf{Synthetic data for segmentation.}
DatasetGAN predicts segmentation masks from GAN representations using limited annotations \cite{zhang2021datasetgan}. 
DiffuMask and Dataset Diffusion generate image-mask pairs \cite{wu2023diffumask,nguyen2023datasetdiffusion}, and latent diffusion and ControlNet support generation conditioned on existing labels \cite{rombach2022ldm,zhang2023controlnet}. 
DGInStyle and ISSA expand appearance variation for driving-scene segmentation \cite{jia2024dginstyle,li2024issa}, and Camouflageator and LAKE-RED generate additional camouflaged examples \cite{he2024camouflageator,zhao2024lakered}. 
These approaches provide additional training images, whose usefulness for training also depends on how the generated content relates to its annotations.

\noindent\textbf{Synthetic data selection and task-based scoring.}
FreeMask filters incorrectly synthesized regions and adjusts sampling across semantic masks \cite{yang2023freemask}.
FLEX-Seg handles synthetic image-mask misalignment during segmentation training \cite{kim2026flexseg}, and Adamkiewicz \etal\ examine the gap between visual realism and synthetic-to-real classification performance \cite{adamkiewicz2026pretty}.
CuRB ranks degradation-type and severity regions using matched short-training probes \cite{he2026hard}. We instead score individual candidates from the same source with a frozen reference and add local repair.
SDS keeps images whose CLIP image-text similarity drops under patch mixing and filters annotations against CLIP activation maps \cite{tang2025sds}.
Zhang \etal\ find that dense scenes and detail-preserving instances train better and propose SENSE, which reassigns pseudo-labels by optimal transport \cite{zhang2026sense}. FLEX-Seg and SENSE act during segmentation training and are complementary to our selection.
Our score is fitted to the task on frozen self-supervised features \cite{oquab2024dinov2}, uses standardized class responses, and measures where blur-induced class responses fall relative to the inherited label.

\noindent\textbf{Local image editing.}
Blended Diffusion edits specified regions with text guidance \cite{avrahami2022blended}, while DiffEdit derives masks by comparing diffusion predictions under different text conditions \cite{couairon2023diffedit}. Latent diffusion supports inpainting, and ControlNet guides structure using conditions such as edges and segmentation maps \cite{rombach2022ldm,zhang2023controlnet}. Our selection score also guides editing: its source-relative discrepancy is attributed to label regions, whose boundaries define the edit extent.

\begin{figure*}[t]
\centering
\setlength{\abovecaptionskip}{0cm}
\setlength{\belowcaptionskip}{-0.1cm}
\includegraphics[width=\textwidth]{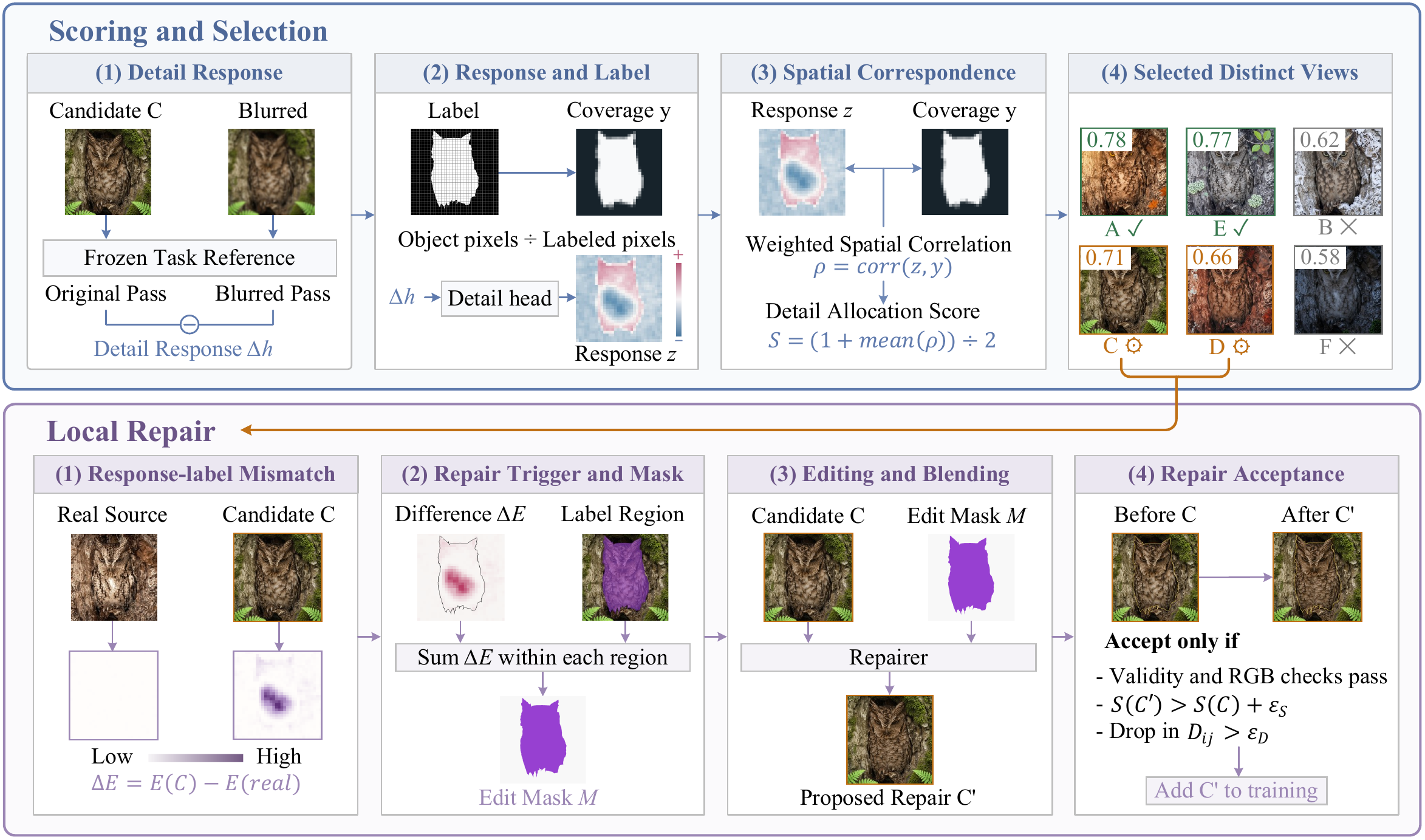}
\caption{\textbf{Detail allocation scoring and local repair.}
Top: Differences between original and blurred reference passes yield
class responses (1), which share a grid with the class fractions of the
inherited label (2).
Weighted spatial correlations between the two define the score (3) for
selecting candidates and repair parents (4).
Bottom: Source-relative discrepancies (1) are aggregated over label
regions to trigger repair and choose edit regions in multiclass tasks (2).
Binary tasks edit the complete foreground.
The generator proposes up to two masked edits (3), and the first proposal
passing the acceptance checks joins the training views (4).}
\label{fig:method}
\end{figure*}

\section{Method}
\label{sec:method}
 
For each source $i$ we have a real photograph $x_i$, its label map $Y_i$, and six generated candidate images $g_{i1},\dots,g_{i6}$ in the same image coordinates. A fixed validity list excludes candidates with known image-label conflicts, and the ignored pixels of $Y_i$ take no part in scoring or repair. From the remaining candidates we keep at most two and consider up to two more for repair (\cref{fig:method}). Selected candidates and accepted repairs retain $Y_i$ for downstream training.

\subsection{Task-Conditioned Detail Responses}
\label{sec:response}

A frozen DINOv2 encoder $\phi$ \cite{oquab2024dinov2} produces a grid of $P$ feature cells for each image, indexed by $t\in\{1,\dots,P\}$.
A \emph{task reference} $f$ processes these features through a mapping $h_f$, comprising a feature adapter, fitted feature normalization, and two spatial convolution layers. 
A coarse head reads $h_f$ of the blurred features and a detail head reads the blur-induced response of \cref{eq:detail}. Their summed logits are fitted on real training images by cross-entropy with the class fractions in each cell (\supp{A}). 
We freeze two references fitted on source-disjoint training subsets. The set $\mathcal F_i$ contains those fitted without source $i$ or its related views. Sources in either subset use only the other reference ($|\mathcal F_i|=1$); sources outside both, as in the salient task, use both references.

For an image $u$ let $B_\sigma u$ be its Gaussian-blurred copy. With token-wise $\ell_2$-normalized features
\begin{equation}
F(u)=\operatorname{norm}\phi(u),\qquad \bar F(u)=\operatorname{norm}\phi(B_\sigma u),
\label{eq:features}
\end{equation}
the blur-induced response of the reference is the change of its representation when detail is attenuated by the blur,
\begin{equation}
\Delta h_f(u,t)=h_f(F(u))_t-h_f(\bar F(u))_t ,
\label{eq:detail}
\end{equation}
computed for the two inputs separately and then subtracted. We use $\sigma=4$ on images resized to a longer side of 512 pixels (\supp{A}).
Let $W_{fc}$ be the weight of class $c$ in the detail head of reference $f$, among $K$ classes. The contrast direction of class $c$ against the other classes is formulated as
\begin{equation}
v_{fc}=W_{fc}-\frac{1}{K-1}\sum_{d\ne c}W_{fd},
\label{eq:contrast}
\end{equation}
and the signed response of class $c$ in cell $t$ is
\begin{equation}
z_{fc}(u,t)=v_{fc}^{\mathsf T}\Delta h_f(u,t).
\label{eq:response}
\end{equation}
A positive blur-induced class response favors class $c$ over the mean of the other classes; contributions shared equally across classes cancel. For binary segmentation, the direction reduces to $W_{f,\mathrm{fg}}-W_{f,\mathrm{bg}}$. These spatial responses define the detail allocation score, while the full prediction from both heads gives the label-agreement baseline (\cref{sec:protocol}).

\subsection{Spatial Correspondence and Candidate Selection}
\label{sec:score}
The inherited label enters as class fractions on the same spatial grid as the response maps. With $m_{itc}$ denoting the number of valid pixels of class $c$ in cell $t$ and $n_{it}=\sum_c m_{itc}$,
\begin{equation}
y_{itc}=\frac{m_{itc}}{n_{it}},\qquad w_{it}=\frac{n_{it}}{\sum_s n_{is}},
\label{eq:labels}
\end{equation}
so a boundary cell keeps the fractions of both classes. Cells without valid labeled pixels are excluded, and the weights sum to one over the remaining cells. For a spatial map $a_t$ we write $\widehat a_t$ for its standardization to weighted mean zero and weighted variance one under $w_{it}$. Let $\mathcal C_i$ be the classes present in $Y_i$ with nonzero spatial variance, fixed by the label and shared by all candidates of the source. The {detail allocation score} is defined as follows:
\begin{equation}
\begin{gathered}
\rho_{ifc}(u)=\sum_t w_{it}\,\widehat z_{fc}(u,t)\,\widehat y_{itc},\\
S_i(u)=\frac12\Bigl[1+\frac{1}{|\mathcal F_i||\mathcal C_i|}\sum_{f\in\mathcal F_i}\sum_{c\in\mathcal C_i}\rho_{ifc}(u)\Bigr].
\end{gathered}
\label{eq:score}
\end{equation}
The score lies in $[0,1]$ and increases as relative class responses align with the inherited label's spatial class fractions. Standardization makes it invariant to positive rescaling or constant shifts of a response map, so response strength alone does not increase it. Eligible classes and task references contribute equally. For binary segmentation, foreground and background correlations are identical, so their average reduces to the foreground term. We rank candidates within each source under its inherited label. Candidates with a spatially constant class response have undefined correlations and are excluded from ranking (\supp{A}).


We select up to two candidates per source. Let $S_{(1)}\ge S_{(2)}$ be the two highest scores among valid, scorable candidates. An admissible pair has sorted scores $S_{ab,(k)}\ge S_{(k)}-\delta$ for $k=1,2$, where $\delta=1/P$. Denoting the admissible pairs by $\mathcal P_i$, we select
\begin{equation}
\begin{gathered}
m_i(a,b)=\min\{d(x_i,g_{ia}),\,d(x_i,g_{ib}),\,d(g_{ia},g_{ib})\},\\
\{a^*,b^*\}=\arg\max_{\{a,b\}\in\mathcal P_i}m_i(a,b),
\end{gathered}
\label{eq:pair}
\end{equation}
where $d$ is the cosine distance between mean DINOv2 features. The score stays the first criterion, and among near-tied candidates the rule avoids two views that repeat the real source image or each other. Ties are resolved by candidate index. A source with one scorable candidate keeps one, and a source with none keeps only its photograph.

\subsection{Two-Step Local Repair}
\label{sec:repair}

\noindent\textbf{Locate.}
The detail allocation score can be expressed through a per-cell discrepancy, formulated as
\begin{equation}
\begin{gathered}
E_i(u,t)=\frac{1}{|\mathcal F_i||\mathcal C_i|}\sum_{f\in\mathcal F_i}\sum_{c\in\mathcal C_i}\bigl(\widehat z_{fc}(u,t)-\widehat y_{itc}\bigr)^2,\\
\sum_t w_{it}\,E_i(u,t)=4\,[1-S_i(u)].
\end{gathered}
\label{eq:local}
\end{equation}
The standardized maps' weighted zero means and unit variances give this equality (\supp{A}). To identify region contributions to the score gap, we attribute the discrepancy difference $\Delta E_{ij}(t)=E_i(g_{ij},t)-E_i(x_i,t)$ to label regions. Let $\mathcal R_i$ partition valid pixels into class regions, $n_{itr}$ count region $r$'s pixels in cell $t$, and $N_i=\sum_t n_{it}$. We define
\begin{equation}
\begin{aligned}
B_{ijr}&=\sum_t\frac{n_{itr}}{N_i}\Delta E_{ij}(t),\\
\sum_{r\in\mathcal R_i}B_{ijr}&=4\bigl[S_i(x_i)-S_i(g_{ij})\bigr].
\end{aligned}
\label{eq:excess}
\end{equation}
Eligible regions are the connected regions of every class in multiclass tasks, including large classes such as road and sky on Cityscapes, and the complete foreground, with its disconnected parts, in binary tasks. We take the eligible region $r^*$ with the largest positive contribution and form $M_{ij}$ from it and the eligible regions that overlap the cell $t^*=\arg\max_{t:\,n_{itr^*}>0}\Delta E_{ij}(t)$. In binary tasks the mask is therefore always the complete foreground, and the contribution decides whether a parent is edited.

We select up to two repair parents, $C$ and $D$, by descending score from valid, scorable candidates outside the selected pair. Each parent is processed independently with one region group. We skip sources without a scorable real image and parents without an eligible region with positive contribution.

\noindent\textbf{Edit and check.}
We reuse each parent's generator and task-specific conditioning, using $M_{ij}$ for local editing. The editor takes the parent as RGB input and is instructed to preserve the labeled layout and surrounding appearance. For degraded salient images, repair retains the original two stages: appearance editing followed by restoration of the parent's degradation condition (\supp{B}). We allow at most two proposals per parent and mask and retain the \mbox{first that passes the checks below.}

We blend the proposal into its parent within $M_{ij}$ to obtain $g'_{ij}$, keeping the remaining pixels and $Y_i$ unchanged. Using the same frozen references, we recompute the score and measure discrepancy over the full edit mask:
\begin{equation}
D_{ij}(u)=\frac{\sum_t n_{it}(M_{ij})E_i(u,t)}{\sum_t n_{it}(M_{ij})},
\label{eq:core}
\end{equation}
where $n_{it}(M_{ij})$ counts the masked valid pixels in cell $t$. A proposal must pass validity checks and meet
\begin{equation}
\begin{aligned}
S_i(g'_{ij}) &> S_i(g_{ij})+\varepsilon_S,\\
D_{ij}(g'_{ij}) &< D_{ij}(g_{ij})-\varepsilon_D.
\end{aligned}
\label{eq:accept}
\end{equation}
The score condition tests the whole image and the discrepancy condition tests the edited region. Both are needed because whole-image standardization lets an edit inside the mask also change the discrepancy outside it. Within-mask RGB changes must also meet minimum magnitude and coverage thresholds at the training resolution (\supp{B}).

\subsection{Training with Source-Level Sampling}
\label{sec:training}
For source $i$, the training bank $\mathcal B_i=\mathcal S_i\cup\mathcal A_i$ contains the selected candidates $\mathcal S_i$ and accepted repairs $\mathcal A_i$, with at most four generated views. Unedited repair parents are not included. Under a fixed source schedule $\pi$, real-image slots use $x_i$ and generated-image slots cycle through $\mathcal B_i$, falling back to $x_i$ if it is empty. Repairs change the frequencies of individual views within existing slots, while source frequencies and the number of updates remain fixed. Images and labels receive aligned spatial augmentations.

\begin{table*}[t]
\setlength{\abovecaptionskip}{0cm}
\setlength{\belowcaptionskip}{-0.3cm}
\centering
\caption{\textbf{Camouflaged object detection with two segmenters.} Both segmenters use the same selected candidates and accepted repairs. Each row, including \BASE{}, reports one training run under our source schedule. Bold marks the best displayed value within each block.}
\label{tab:cod}
\resizebox{\textwidth}{!}{%
\setlength{\tabcolsep}{1.3mm}
\begin{tabular}{l|c|cccc|cccc|cccc|cccc}
\toprule
\multicolumn{1}{c|}{} & \multicolumn{1}{c|}{} & \multicolumn{4}{c|}{\textit{CHAMELEON}} & \multicolumn{4}{c|}{\textit{CAMO}} & \multicolumn{4}{c|}{\textit{COD10K}} & \multicolumn{4}{c}{\textit{NC4K}} \\ \cline{3-18}
\multicolumn{1}{l|}{\multirow{-2}{*}{Strategy}} & \multicolumn{1}{c|}{\multirow{-2}{*}{Encoder}} & $M\downarrow$ & $F_\beta\uparrow$ & $E_\phi\uparrow$ & $S_\alpha\uparrow$ & $M\downarrow$ & $F_\beta\uparrow$ & $E_\phi\uparrow$ & $S_\alpha\uparrow$ & $M\downarrow$ & $F_\beta\uparrow$ & $E_\phi\uparrow$ & $S_\alpha\uparrow$ & $M\downarrow$ & $F_\beta\uparrow$ & $E_\phi\uparrow$ & $S_\alpha\uparrow$ \\
\midrule
\multicolumn{18}{l}{\emph{FEDER, 70 epochs}} \\
\BASE & ResNet-50 & 0.029 & 0.836 & 0.941 & 0.890 & 0.085 & 0.746 & 0.837 & 0.779 & 0.038 & 0.694 & 0.882 & 0.806 & 0.051 & 0.785 & 0.892 & 0.833 \\
\ALL & ResNet-50 & 0.033 & 0.824 & 0.925 & 0.883 & 0.082 & 0.746 & 0.841 & 0.783 & 0.038 & 0.691 & 0.881 & 0.804 & 0.051 & 0.788 & 0.890 & 0.834 \\
\RAND & ResNet-50 & 0.033 & 0.824 & 0.937 & 0.883 & 0.080 & 0.753 & 0.843 & 0.787 & 0.038 & 0.686 & 0.878 & 0.804 & 0.050 & 0.786 & 0.893 & 0.835 \\
\AGREE & ResNet-50 & 0.032 & 0.825 & 0.929 & 0.883 & 0.081 & 0.747 & 0.842 & 0.783 & 0.038 & 0.692 & 0.880 & 0.806 & 0.050 & 0.788 & 0.892 & 0.835 \\
\rowcolor{rowhl} \SELECT & ResNet-50 & 0.026 & 0.844 & 0.946 & 0.896 & 0.079 & 0.766 & 0.858 & 0.800 & 0.036 & 0.707 & 0.890 & 0.817 & 0.048 & 0.795 & 0.899 & 0.843 \\
\rowcolor{rowhl} \SELECTR & ResNet-50 & \textbf{0.025} & \textbf{0.851} & \textbf{0.948} & \textbf{0.899} & \textbf{0.074} & \textbf{0.779} & \textbf{0.863} & \textbf{0.807} & \textbf{0.034} & \textbf{0.721} & \textbf{0.894} & \textbf{0.820} & \textbf{0.046} & \textbf{0.806} & \textbf{0.902} & \textbf{0.846} \\
\midrule
\multicolumn{18}{l}{\emph{RUN, 70 epochs}} \\
\BASE & ResNet-50 & 0.026 & 0.858 & 0.947 & 0.897 & 0.071 & 0.783 & 0.868 & 0.805 & 0.031 & 0.744 & 0.904 & 0.827 & 0.042 & 0.826 & 0.910 & 0.851 \\
\ALL & ResNet-50 & 0.026 & 0.856 & 0.945 & 0.896 & 0.071 & 0.777 & 0.863 & 0.797 & 0.031 & 0.742 & 0.903 & 0.827 & 0.043 & 0.822 & 0.910 & 0.852 \\
\RAND & ResNet-50 & 0.028 & 0.855 & 0.946 & 0.893 & 0.071 & 0.778 & \textbf{0.875} & 0.805 & 0.031 & 0.739 & 0.901 & 0.826 & 0.042 & 0.822 & 0.912 & 0.852 \\
\AGREE & ResNet-50 & 0.026 & 0.858 & 0.946 & 0.898 & 0.071 & 0.782 & 0.868 & 0.805 & 0.031 & 0.743 & 0.904 & 0.827 & 0.042 & 0.825 & 0.911 & 0.852 \\
\rowcolor{rowhl} \SELECT & ResNet-50 & 0.026 & 0.860 & \textbf{0.958} & 0.901 & 0.071 & 0.787 & 0.871 & 0.810 & \textbf{0.029} & 0.760 & \textbf{0.912} & 0.840 & 0.042 & 0.828 & \textbf{0.913} & 0.855 \\
\rowcolor{rowhl} \SELECTR & ResNet-50 & \textbf{0.025} & \textbf{0.861} & 0.956 & \textbf{0.906} & \textbf{0.070} & \textbf{0.789} & 0.871 & \textbf{0.816} & \textbf{0.029} & \textbf{0.761} & \textbf{0.912} & \textbf{0.846} & \textbf{0.041} & \textbf{0.832} & \textbf{0.913} & \textbf{0.861} \\
\bottomrule
\end{tabular}}
\vspace{-1mm}
\end{table*}

\begin{table*}[t]
\setlength{\abovecaptionskip}{0cm}
\centering
\caption{\textbf{Degraded salient object detection with CSSAM} under the CSOD10K protocol. WXSOD-real provides cross-dataset evaluation. We follow the CSSAM metrics. \RANDM{} redraws the candidates of each source at random under the condition counts of \SELECT{}. }
\label{tab:sod}
\begin{subtable}[t]{0.485\textwidth}
\setlength{\abovecaptionskip}{0cm}
\centering
\caption{CSOD10K}
\label{tab:sod-csod}
\tablefont
\setlength{\tabcolsep}{3pt}
\renewcommand{\arraystretch}{1.08}
\begin{tabular}{L{70pt}*{4}{C{\dimexpr(\linewidth-100pt)/4\relax}}}
\toprule
Strategy & $M\downarrow$ & $F_\beta^{\max}\uparrow$ & $E_\phi\uparrow$ & $S_\alpha\uparrow$ \\
\midrule
\BASE & 0.035 & 0.870 & 0.916 & 0.888 \\
\ALL & 0.035 & 0.869 & 0.914 & 0.887 \\
\RAND & 0.036 & 0.871 & 0.914 & 0.886 \\
\RANDM & 0.035 & 0.868 & 0.912 & 0.887 \\
\AGREE & 0.035 & 0.870 & 0.915 & 0.889 \\
\rowcolor{rowhl} \SELECT & 0.031 & 0.886 & 0.925 & 0.899 \\
\rowcolor{rowhl} \SELECTR & \textbf{0.030} & \textbf{0.889} & \textbf{0.928} & \textbf{0.901} \\
\bottomrule
\end{tabular}
\end{subtable}\hfill
\begin{subtable}[t]{0.485\textwidth}
\setlength{\abovecaptionskip}{0cm}
\centering
\caption{WXSOD-real}
\label{tab:sod-wxsod}
\tablefont
\setlength{\tabcolsep}{3pt}
\renewcommand{\arraystretch}{1.08}
\begin{tabular}{L{70pt}*{4}{C{\dimexpr(\linewidth-100pt)/4\relax}}}
\toprule
Strategy & $M\downarrow$ & $F_\beta^{\max}\uparrow$ & $E_\phi\uparrow$ & $S_\alpha\uparrow$ \\
\midrule
\BASE & 0.015 & 0.912 & 0.954 & 0.933 \\
\ALL & 0.014 & 0.917 & 0.954 & 0.935 \\
\RAND & 0.014 & 0.916 & 0.958 & 0.936 \\
\RANDM & 0.014 & 0.916 & 0.958 & 0.937 \\
\AGREE & 0.014 & 0.917 & 0.955 & 0.935 \\
\rowcolor{rowhl} \SELECT & \textbf{0.013} & 0.920 & 0.961 & 0.939 \\
\rowcolor{rowhl} \SELECTR & \textbf{0.013} & \textbf{0.924} & \textbf{0.963} & \textbf{0.942} \\
\bottomrule
\end{tabular}
\end{subtable}
\end{table*}

\section{Experiments}
\label{sec:experiments}

\subsection{Implementation Details}
\label{sec:protocol}

\noindent\textbf{Datasets and models.}
COD uses FEDER \cite{He2023Camouflaged} and RUN \cite{he2025run}, trained for 70 epochs on 4,040 labeled CAMO-train and COD10K-train images \cite{le2019camo,fan2020cod10k} and tested on CHAMELEON \cite{skurowski2018chameleon}, CAMO, COD10K, and NC4K \cite{lv2021nc4k}.
Both share selected candidates and accepted repairs; ablations and mechanism analysis use FEDER. Both segmenters use RUN's 70-epoch schedule so that all COD comparisons share a matched budget, and each strategy is compared with its segmenter's retrained real-only baseline (\supp{B}). RUN reproduces its published results in this setting. \supp{C} repeats the FEDER comparison under its own 160-epoch recipe. Degraded SOD uses CSSAM under the CSOD10K protocol \cite{cong2026csod}, with WXSOD-real \cite{chen2025wxsod} for cross-dataset evaluation. Semantic segmentation follows DGInStyle \cite{jia2024dginstyle} on Cityscapes \cite{cordts2016cityscapes} with DAFormer (MiT-B2) \cite{hoyer2022daformer}, evaluated on ACDC \cite{sakaridis2021acdc}, Dark Zurich \cite{sakaridis2019darkzurich}, and Cityscapes validation. Each source has six candidate slots. COD and SOD edit appearance with Qwen-Image-Edit \cite{wu2025qwenimage}, given the photograph and its label mask as reference images. SOD uses 3,000 clean DUTS-TR images \cite{wang2017duts} and then adds low light or backlight and, rarely, fog or rain. Cityscapes uses DGInStyle-based generation with semantic and edge conditioning. We fit two references per task on source-disjoint real training subsets using DINOv2 ViT-S/14 features; \cref{sec:further} tests a CLIP reference for COD. All strategies share validity lists fixed before scoring by an automatic gate, which rejects a candidate when a class loses support relative to the real source (\supp{B}). Repair validity checks automatically verify decoding at the parent's image size and unchanged pixels outside the mask.

{\tightpara
\noindent\textbf{Strategies and training.}
Within a task, all strategies share the source schedule and the number of updates. \BASE{} fills the generated slots with augmented real images. \ALL{} uses all valid candidates, and \RAND{} draws as many valid, scorable candidates per source as \SELECT{}. \AGREE{} keeps the same quotas and ranks candidates by agreement between the inherited label and the references' full prediction (foreground IoU, or mIoU over present classes on Cityscapes). \SELECT{} uses the proposed score, and \SELECTR{} adds its accepted repairs.\par}

{\looseness=-1
\noindent\textbf{Evaluation.}
COD follows FEDER and RUN with MAE $M$, adaptive F-measure $F_\beta$, mean E-measure $E_\phi$ \cite{fan2018emeasure}, and S-measure $S_\alpha$ \cite{fan2017smeasure}. The salient task follows CSSAM with $M$, maximum F-measure $F_\beta^{\max}$, $E_\phi$, and $S_\alpha$, and semantic segmentation reports mIoU (\supp{B}). We report $\dsel=U(\SELECT)-U(\RAND)$ and $\drep=U(\SELECTR)-U(\SELECT)$, where $U$ is downstream test performance. The first reflects only which candidates are kept, and the second the added repairs.\par}

\begin{figure*}[t!]
\setlength{\abovecaptionskip}{1mm}
  \centering
  \includegraphics[width=\linewidth]{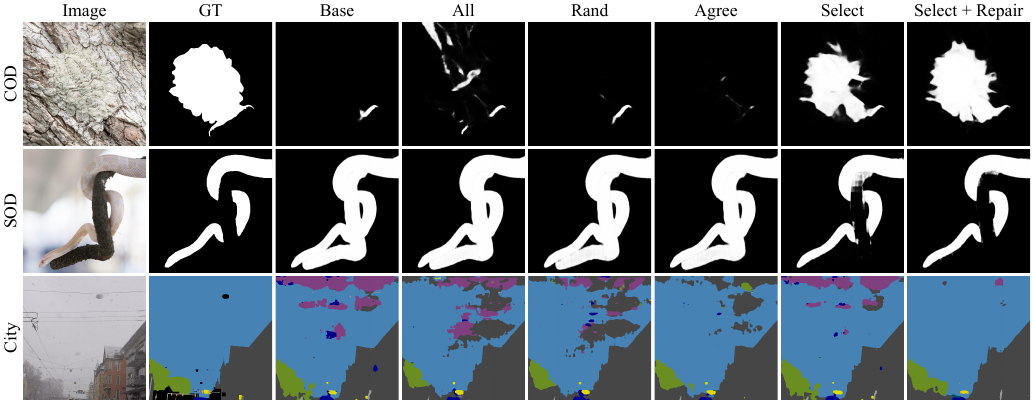}
\caption{\textbf{Qualitative comparison on the three tasks.} Each row shows a test image, its label, and the predictions of the segmenter trained with \BASE{}, \ALL{}, \RAND{}, \AGREE{}, \SELECT{}, and \SELECTR{}. Top: COD. Middle: degraded SOD. Bottom: Semantic segmentation.}
\label{fig:qual}
\end{figure*}

\subsection{Comparison Across Segmentation Tasks}
\label{sec:applications}

On COD10K (\cref{tab:cod,fig:qual}), selection improves $S_\alpha$ over \RAND{} by 0.013 with FEDER and 0.014 with RUN; repair adds 0.003 and 0.006 over \SELECT{}, respectively. Selection also raises $F_\beta$ by 0.021 over \RAND{} with both segmenters. Both steps improve $S_\alpha$ and $F_\beta$ on all four test sets with both segmenters. \ALL{}, \RAND{}, and \AGREE{} remain within 0.002 $S_\alpha$ of the real-only baseline \BASE{} on COD10K.

On CSOD10K (\cref{tab:sod,fig:qual}), selection improves $S_\alpha$ by 0.013 and $F_\beta^{\max}$ by 0.015 over \RAND{}, and repair adds 0.002 and 0.003 over \SELECT{}, respectively. Both steps also improve $S_\alpha$ on WXSOD-real in cross-dataset evaluation.

On ACDC (\cref{tab:sem,fig:qual}), \ALL{}, \RAND{}, and \AGREE{} fall 3.2 to 4.0 mIoU below \BASE{}. \SELECT{} exceeds \RAND{} by 4.261 and \BASE{} by 0.286, and repair adds 2.572. Selection improves mIoU on Dark Zurich but trails \ALL{}, \RAND{}, and \AGREE{} on Cityscapes validation. \SELECTR{} performs best among these strategies on all three evaluations.

\cref{tab:selection} compares the score with existing selection methods on COD10K under the same pool and quota. MUSIQ \cite{ke2021musiq} ranks candidates by image quality, SDS \cite{tang2025sds} combines CLIP-based image and annotation reliability, and \AGREE{} uses the full predictions of the same task references. \AGREE{} is the strongest baseline, and \SELECT{} exceeds it by 0.011 in $S_\alpha$ and 0.015 in $F_\beta$ at lower $M$ and higher $E_\phi$. No baseline shares more than 11.8\% of its pairs with \SELECT{}.

\begin{table}[t]
\setlength{\abovecaptionskip}{0cm}
\centering
\caption{\textbf{Adverse-condition semantic segmentation} (DAFormer, MiT-B2, mIoU). ACDC reports the macro mean across conditions, and Cityscapes validation assesses source-domain performance.}
\label{tab:sem}
\tablefont
\setlength{\tabcolsep}{2pt}
\renewcommand{\arraystretch}{1.08}
\begin{tabular}{L{62pt}*{3}{C{\dimexpr(\linewidth-78pt)/3\relax}}}
\toprule
Strategy & ACDC$\uparrow$ & \makecell{Dark\\Zurich$\uparrow$} & \makecell{Cityscapes\\val$\uparrow$} \\
\midrule
\BASE & 46.029 & 21.883 & 72.738 \\
\ALL & 42.013 & 19.653 & 73.415 \\
\RAND & 42.054 & 19.602 & 73.261 \\
\AGREE & 42.789 & 20.149 & 73.129 \\
\rowcolor{rowhl} \SELECT & 46.315 & 22.220 & 72.873 \\
\rowcolor{rowhl} \SELECTR & \textbf{48.887} & \textbf{23.488} & \textbf{75.070} \\
\bottomrule
\end{tabular}\vspace{-1mm}
\end{table}

\begin{table}[t]
\setlength{\abovecaptionskip}{0cm}
\centering
\caption{\textbf{Selection baselines (top) and ablations (bottom) on COD10K} (FEDER) under the same pool and per-source quota. Response strength ranks by magnitude alone. Same pair is the \% of sources matching \SELECT{}'s pair.}
\label{tab:selection}
\resizebox{\columnwidth}{!}{%
\setlength{\tabcolsep}{1.6mm}
\begin{tabular}{l|cccc|c}
\toprule
Selection & $M\downarrow$ & $F_\beta\uparrow$ & $E_\phi\uparrow$ & $S_\alpha\uparrow$ & \makecell{same\\pair} \\
\midrule
\RAND & \textbf{0.038} & 0.686 & 0.878 & 0.804 & 6.2\% \\
\textsc{musiq} & \textbf{0.038} & 0.690 & 0.879 & \textbf{0.806} & 8.4\% \\
\textsc{sds} & \textbf{0.038} & 0.689 & \textbf{0.880} & 0.805 & 7.1\% \\
\AGREE & \textbf{0.038} & \textbf{0.692} & \textbf{0.880} & \textbf{0.806} & 11.8\% \\
\midrule
response strength & 0.039 & 0.685 & 0.877 & 0.807 & 21.6\% \\
full-response correlation & 0.038 & 0.695 & 0.881 & 0.813 & 15.9\% \\
plain top-two & 0.037 & 0.699 & 0.883 & 0.814 & 92.6\% \\
\rowcolor{rowhl} \SELECT & \textbf{0.036} & \textbf{0.707} & \textbf{0.890} & \textbf{0.817} & 100\% \\
\bottomrule
\end{tabular}%
}
\vspace{-2mm}
\end{table}

\subsection{Ablation Studies}
\label{sec:ablation}

\noindent\textbf{Blur-induced response.}
We replace blur-induced responses with the full prediction's class-contrast logits, keeping the fitted references, pool, spatial correlation, and pair-selection rule fixed. Full-response correlation gains 0.007 $S_\alpha$ over \AGREE{}, while \SELECT{} adds another 0.004 $S_\alpha$ and 0.012 $F_\beta$ (\cref{tab:selection}). \cref{tab:seeds} repeats this contrast over three seeds.

\noindent\textbf{Response strength.}
Using the same references, we select each source's top two candidates by the mean magnitude of unnormalized class-contrast responses (\supp{B}). On COD10K with FEDER, this gives 0.807 $S_\alpha$ versus 0.814 for plain top-two, our score without the near-tie rule (\cref{tab:selection}), so magnitude alone does not explain the gain.

\begin{table}[t]
\setlength{\abovecaptionskip}{0cm}
\centering
\caption{\textbf{Repair ablations} (COD10K, FEDER). \PARENT{} adds the unedited parents of \SELECTR{}'s accepted repairs. \RANDR{} shifts each edit mask to a random valid location. Edit area is the mean masked image fraction.}
\label{tab:repair}
\resizebox{\columnwidth}{!}{%
\setlength{\tabcolsep}{1.3mm}
\begin{tabular}{l|cc|cccc}
\toprule
Strategy & \makecell{added\\views} & \makecell{edit\\area} & $M\downarrow$ & $F_\beta\uparrow$ & $E_\phi\uparrow$ & $S_\alpha\uparrow$ \\
\midrule
\SELECT & 0 & 0 & 0.036 & 0.707 & 0.890 & 0.817 \\
\PARENT & 1,858 & 0 & 0.036 & 0.714 & 0.891 & 0.818 \\
\makecell[l]{\textsc{select+}\\\textsc{random-repair}} & 699 & 10.8\% & 0.035 & 0.716 & 0.892 & 0.818 \\
\rowcolor{rowhl} \SELECTR & 1,858 & 10.8\% & \textbf{0.034} & \textbf{0.721} & \textbf{0.894} & \textbf{0.820} \\
\bottomrule
\end{tabular}%
}
\vspace{-1mm}
\end{table}

\begin{figure*}[t]
\centering
\includegraphics[width=1\textwidth]{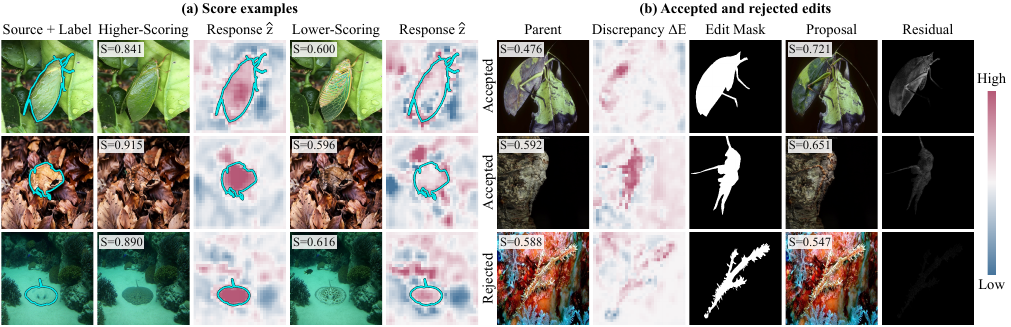}
\caption{\looseness=-1 \textbf{Score and repair examples on COD.} (a) Three real sources with inherited labels, higher- and lower-scoring candidates, and their standardized response maps $\widehat z$. Red and blue mark responses above and below each image's mean on a shared diverging scale. (b) Two accepted edits and one rejected edit, showing each parent, source-relative discrepancy map, edit mask, proposal, and residual between parent and proposal. Source-relative discrepancy maps share a color scale. The rejected edit lowers the score from 0.588 to 0.547.}
\label{fig:examples}
\end{figure*}

\noindent\textbf{Local repair.}
\cref{tab:repair} compares repair controls within the same training slots. \PARENT{} adds the unedited parents of accepted repairs, so sources and added views match. \SELECTR{} exceeds it by 0.002 $S_\alpha$ and 0.007 $F_\beta$, and \cref{tab:seeds} repeats the contrast over three seeds.
\RANDR{} shifts masks to random valid locations (\supp{B}). It yields 699 accepted repairs versus 1,858 for \SELECTR{} and performs worse on all four metrics. Other trigger rules perform similarly on COD (\supp{C}).

\begin{figure}[t]
\centering
\includegraphics[width=\linewidth]{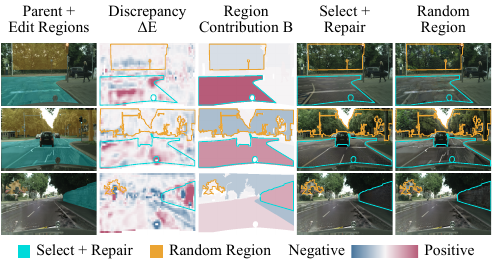}
\caption{\textbf{Region choice on Cityscapes.} Three parents with the regions edited by \SELECTR{} (cyan) and \RANDG{} (orange), the source-relative discrepancy map, the region contributions, and the two repaired images.}
\label{fig:csregions}
\end{figure}

\noindent\textbf{Region choice on Cityscapes.}
We test region choice on Cityscapes, since only the multiclass task chooses among label regions. \cref{tab:semrepair} compares two controls. \PARENT{} adds the unedited parents of accepted repairs. \RANDG{} starts from \SELECTR{}'s attempted parents and randomly chooses each anchor among eligible regions whose mask area is closest to \SELECTR{}'s. The mask rule, proposal limit, and acceptance checks stay fixed (\supp{B}). On ACDC, \PARENT{} and \RANDG{} reach 46.945 and 47.665 mIoU versus 48.887 for \SELECTR{}, so edits beat unedited parents, and guided regions beat random ones. \cref{fig:csregions} shows parents where \SELECTR{} and \RANDG{} edit different regions.

\begin{table}[t]
\setlength{\abovecaptionskip}{0cm}
\centering
\caption{\textbf{Repair controls on the semantic task} (DAFormer, mIoU). \PARENT{} adds the unedited parents of accepted repairs. \RANDG{} draws a random anchor region of comparable area on the same attempted parents.}
\label{tab:semrepair}
\tablefont
\setlength{\tabcolsep}{2pt}
\renewcommand{\arraystretch}{1.08}
\begin{tabular}{L{70pt}C{24pt}*{3}{C{\dimexpr(\linewidth-114pt)/3\relax}}}
\toprule
Strategy & \makecell{added\\views} & ACDC$\uparrow$ & \makecell{Dark\\Zurich$\uparrow$} & \makecell{Cityscapes\\val$\uparrow$} \\
\midrule
\SELECT & 0 & 46.315 & 22.220 & 72.873 \\
\PARENT & 3,427 & 46.945 & 22.996 & 74.625 \\
\makecell[l]{\textsc{select+}\\\textsc{random-region}} & 3,169 & 47.665 & 23.151 & 74.617 \\
\rowcolor{rowhl} \SELECTR & 3,427 & \textbf{48.887} & \textbf{23.488} & \textbf{75.070} \\
\bottomrule
\end{tabular}
\end{table}

\subsection{Mechanism Analysis}
\label{sec:mechanism}

\noindent\textbf{Same-source candidate sets.}
From each source with at least two valid, scorable candidates, we take the highest- and lowest-scoring ones. We remove sources from both sets until standardized mean differences (SMD) in MUSIQ, label agreement, and source similarity satisfy $|\mathrm{SMD}|\le 0.07$. A random set shares the retained sources. All three sets train FEDER on synthetic images only (\supp{B}). The high-score set exceeds the low-score set by 0.017 $F_\beta$, with the random set between them on every metric (\cref{tab:controlled}).

\begin{table}[t]
\setlength{\abovecaptionskip}{0cm}
\centering
\caption{\textbf{Same-source candidate sets on COD10K} (FEDER, synthetic-only). Matching keeps 96.5\% of sources and balances MUSIQ, label agreement, and source similarity across the two sets ($|\mathrm{SMD}|\le 0.07$). Score is the set mean.}
\label{tab:controlled}
\tablefont
\setlength{\tabcolsep}{2pt}
\renewcommand{\arraystretch}{1.08}
\begin{tabular}{L{87pt}*{5}{C{\dimexpr(\linewidth-111pt)/5\relax}}}
\toprule
Training set & score & $M\downarrow$ & $F_\beta\uparrow$ & $E_\phi\uparrow$ & $S_\alpha\uparrow$ \\
\midrule
highest-score candidate & 0.72 & \textbf{0.052} & \textbf{0.614} & \textbf{0.830} & \textbf{0.768} \\
random candidate & 0.68 & 0.053 & 0.611 & \textbf{0.830} & 0.766 \\
lowest-score candidate & 0.64 & 0.055 & 0.597 & 0.829 & 0.765 \\
\bottomrule
\end{tabular}\vspace{-1mm}
\end{table}

\noindent\textbf{Label agreement after repair.}
Acceptance and repair guidance share a reference, so a real-only segmenter that has not seen the source checks each accepted repair and its parent. Inside COD edit masks, which cover the labeled foreground, its agreement with the inherited label is 2.4 points higher for repairs than for parents (\supp{C}).

\noindent\textbf{What the score and the edit see.}
\cref{fig:examples}~(a) contrasts candidates with similar label agreement but different allocation scores. Positive responses fall on the labeled objects in higher-scoring candidates and around outlines or in the background in lower-scoring ones. \cref{fig:examples}~(b) shows accepted and rejected repairs with their discrepancy maps and masks. Region contributions decide whether a parent is edited, the inherited label supplies the mask, and pixels outside it stay unchanged. Repair counts are in \supp{C}.

\begin{figure}[t]
\centering
\includegraphics[width=\linewidth]{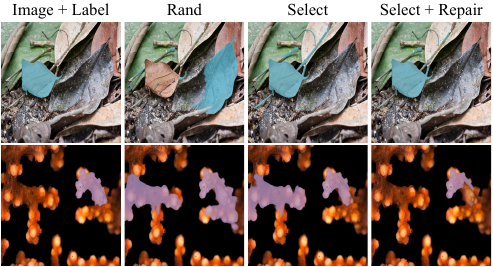}
\caption{\looseness=-1 \textbf{Predictions on COD10K} by FEDER trained with \RAND{}, \SELECT{}, and \SELECTR{}. Rows show two images from the difficult subsets of \cref{tab:diff}, alongside their predicted masks.}\vspace{-1mm}
\label{fig:predictions}
\end{figure}

\noindent\textbf{Where the gains fall.}
\cref{tab:diff} reports COD10K $F_\beta$ by difficulty. Selection improves over \RAND{} on every subset, by about 0.02 except on indefinable boundaries. Repair adds 0.014 to 0.019 on the three difficult subsets but only 0.005 on the remaining images (\cref{fig:predictions}).
For degraded SOD, the valid pool is about half low light and half backlight with under 1\% fog and rain, so the data mainly cover illumination changes. Selection raises the low-light share to 63.65\%, and the selected low-light images are not milder in mean recorded strength. \RANDM{} redraws each source's candidates at random to match \SELECT{}'s condition counts (\cref{tab:sod}, \supp{B}). It reaches 0.887 $S_\alpha$ on CSOD10K versus 0.886 for \RAND{} and 0.899 for \SELECT{}, {so matching the condition mix does not recover the selection gain}.
\begin{table}[t]
\setlength{\abovecaptionskip}{0cm}
\centering
\caption{\textbf{Gains by difficulty on COD10K} (FEDER, $F_\beta$). Results use the checkpoints of \cref{tab:cod}. Difficulty attributes may overlap.}
\label{tab:diff}
\tablefont
\setlength{\tabcolsep}{2pt}
\renewcommand{\arraystretch}{1.30}
\begin{tabular}{L{78pt}C{27pt}C{25pt}C{29pt}C{\dimexpr\linewidth-179pt\relax}}
\toprule
Test images & Number & \RAND & \SELECT & \SELECTR \\
\midrule
small object & 1,379 & 0.613 & 0.635 & \textbf{0.654} \\
occlusion & 928 & 0.657 & 0.678 & \textbf{0.692} \\
indefinable boundaries & 610 & 0.694 & 0.704 & \textbf{0.720} \\
none of the three & 252 & 0.877 & 0.897 & \textbf{0.902} \\
\midrule
all & 2,026 & 0.686 & 0.707 & \textbf{0.721} \\
\bottomrule
\end{tabular} \vspace{-1mm}
\end{table}

\begin{table}[t]
\setlength{\abovecaptionskip}{0cm}
\centering
\caption{\tightpara \textbf{Reference encoders on COD10K} (FEDER, S-measure). References are refitted and candidates rescored per encoder.}
\label{tab:encoder}
\tablefont
\setlength{\tabcolsep}{2pt}
\renewcommand{\arraystretch}{1.30}
\begin{tabular}{L{86pt}*{4}{C{\dimexpr(\linewidth-106pt)/4\relax}}}
\toprule
Reference encoder & \RAND & \AGREE & \SELECT & $\dsel$ \\
\midrule
DINOv2 ViT-S/14 & 0.804 & 0.806 & \textbf{0.817} & +0.013 \\
CLIP ViT-B/16 & 0.804 & 0.805 & \textbf{0.811} & +0.007 \\
\bottomrule
\end{tabular}\vspace{-1mm}
\end{table}

\subsection{Further Analysis}
\label{sec:further}

\noindent\textbf{Training seeds.}
Earlier tables report single training runs. \cref{tab:seeds} repeats the main COD contrasts with FEDER over three seeds, keeping the pool, references, selection lists, and accepted repairs fixed. Strategies share initialization, source schedules, and augmentations within each seed (\supp{C}). Selection gains exceed the seed variation, and repair gains over unedited parents are small but consistent. \supp{C} repeats the semantic comparison over three seeds.

\noindent\textbf{Reference encoder.}
Refitting the COD references on frozen CLIP ViT-B/16 features \cite{radford2021clip} and retraining FEDER (\supp{B}) gives \SELECT{} a gain of 0.007 $S_\alpha$ over \RAND{}, versus 0.013 with DINOv2 (\cref{tab:encoder}).

\noindent\textbf{Cost.}
On an existing COD pool with fitted references, \SELECT{} costs 0.24 GPU hours and \SELECTR{} an estimated 4.70, including rejected proposals. Generating the shared pool costs an estimated 11.65 GPU hours (\supp{C}). Selection needs only two feature passes per candidate (original and blurred) and no downstream training.

\begin{table}[t]
\setlength{\abovecaptionskip}{0cm}
\centering
\caption{\textbf{Paired differences over three training seeds on COD10K} (FEDER), as mean and standard deviation. Data lists are fixed across seeds. Repair policies omit the \textsc{select+} prefix.}
\label{tab:seeds}
\tablefont
\setlength{\tabcolsep}{2pt}
\renewcommand{\arraystretch}{1.08}
\begin{tabular}{L{112pt}*{2}{C{\dimexpr(\linewidth-124pt)/2\relax}}}
\toprule
Contrast & $\Delta S_\alpha$ & $\Delta F_\beta$ \\
\midrule
\SELECT{} $-$ \RAND & 0.011 $\pm$ 0.002 & 0.013 $\pm$ 0.007 \\
\SELECT{} $-$ \AGREE & 0.010 $\pm$ 0.001 & 0.011 $\pm$ 0.004 \\
\SELECT{} $-$ full-response & 0.004 $\pm$ 0.002 & 0.012 $\pm$ 0.003 \\
\midrule
\textsc{repair} $-$ \textsc{parent} & 0.003 $\pm$ 0.001 & 0.007 $\pm$ 0.001 \\
\textsc{repair} $-$ \SELECT{} & 0.005 $\pm$ 0.001 & 0.018 $\pm$ 0.003 \\
\textsc{repair} $-$ \BASE & 0.016 $\pm$ 0.002 & 0.030 $\pm$ 0.002 \\
\bottomrule
\end{tabular}\vspace{-1mm}
\end{table}

\section{Discussion and Limitations}
\label{sec:discussion}

Selection and repair contribute differently across tasks. On COD10K and CSOD10K, selection accounts for most of the $S_\alpha$ improvement. On ACDC, selection mainly recovers the loss from using all candidates, and repair provides most of the gain over real-only training. The semantic task is also the only one where the region contributions choose among label regions (\cref{tab:semrepair}). In the binary tasks every edit covers the complete foreground, so they only decide which parents are edited. 
\ALL{} and the equal-quota \RAND{} and \AGREE{} stay within 0.002 $S_\alpha$ of \BASE{} on COD10K and CSOD10K and fall below it on ACDC,
so the selection gain comes from which candidates enter training and not from how many.

\noindent\textbf{Limitations.}
TCDA requires a real source image for each inherited label and a task reference fitted on real training images. The source provides the reference discrepancy, and the label defines the boundaries of eligible edit regions. The pretraining data of public generators are not fully documented, so overlap between generated images and test images cannot be ruled out. This limitation is shared by every method that trains on images from public generators.

\section{Conclusion}
\label{sec:conclusion}

We studied whether the detail of a synthetic segmentation image supports each class where its inherited label places that class. A task reference fitted on real training images turns this into a detail allocation score. The score ranks candidates, and its source-relative region contributions guide local repair within complete label regions without changing the annotations or downstream segmenter. Across three tasks, selection outperforms random and label-agreement subsets, and local repair adds further gains. The score needs only a frozen encoder and a small task reference, so it can screen an existing pool without downstream training. Future work will use the region contributions to guide generation itself.

\newpage

\section*{LLM Usage Statement}
Large language models (Claude and ChatGPT) were used to polish the wording of all sections and to help debug code and write plotting scripts. The authors designed and ran all experiments, analyzed all results, and take full \mbox{responsibility for all content of this paper.}

{
    \small
    \bibliographystyle{ieeenat_fullname}
    \bibliography{main}
}

\end{document}